%% file: main.tex
\documentclass{opt2021}

\usepackage{verbatim}
\usepackage{inconsolata}
\usepackage{listings}
\usepackage{cleveref}
\usepackage{wrapfig}

\include{commands}

\newcommand{\papertitle}{Adaptive Optimization with Examplewise Gradients}
\newcommand{\runningpapertitle}{\papertitle}
\title[\runningpapertitle]{\papertitle}

\optauthor{%
\Name{Julius Kunze} \Email{juliuskunze@gmail.com}\\
\Name{James Townsend} \Email{james.townsend@cs.ucl.ac.uk}\\
\Name{David Barber} \Email{david.barber@cs.ucl.ac.uk}\\
\addr 
Department of Computer Science \\
University College London}

\begin{document}

\maketitle

\begin{abstract}
We propose a new, more general approach to the design of stochastic gradient-based optimization methods for machine learning. In this new framework, optimizers assume access to a \emph{batch} of gradient estimates per iteration, rather than a single estimate. This better reflects the information that is actually available in typical machine learning setups. To demonstrate the usefulness of this generalized approach, we develop Eve, an adaptation of the Adam optimizer which uses examplewise gradients to obtain more accurate second-moment estimates. We provide preliminary experiments, without hyperparameter tuning, which show that the new optimizer slightly outperforms Adam on a small scale benchmark and performs the same or worse on larger scale benchmarks. Further work is needed to refine the algorithm and tune hyperparameters. 
\end{abstract}

\section{Introduction}

In many practical settings in machine learning, we would like to solve a problem of the form
\begin{equation}\label{eq:ml-prob}
    \argmin_\param \mathbb{E}[\objective(\data, \param)].
\end{equation}
We do not know the underlying distribution of \(\data\), but we do have access to a (often large) set of samples \(X = (\data_1, \ldots, \data_D)\) from the distribution. In this situation, we optimize the approximate loss function
\begin{equation}
    L(\param):=\frac{1}{D}\sum_d \objective(\data_d, \param)\approx\mathbb{E}[\objective(\data, \param)], 
\end{equation}
using stochastic gradient estimates of the form
\begin{equation}\label{eq:g-hat}
    \batchg(\param):=\frac{1}{B}\sum_{\batchi=1}^B \grad_\param \objective(x_{d_b}, \param),
\end{equation}
where \(d_1,\ldots,d_B\) is a random sub-sample of the index set \(\{1,\ldots, D\}\), drawn at each iteration of an optimization process. We refer to \(B\) as the `batch size'.

Optimization methods based on stochastic gradients typically assume access to the estimator \(\batchg\), without making use of the `examplewise' gradient estimates \(\grad_\param \objective(x_{d_b}, \param)\). For optimizers whose iterates depend only linearly on \(\batchg\), this makes sense---the empirical mean over the batch is then no more informative than the examplewise estimates. However, state-of-the-art methods, such as the Adam optimizer \citep{kingma2014adam}, typically have nonlinear dependence on \(\batchg\), and in these situations it may be beneficial to make direct use of the examplewise estimates.
Especially at the beginning of training, Adam's second-moment estimates can be unstable, which can harm learning \cite{liu2019variance}.

Our contribution is twofold. Firstly, we propose \emph{Eve}, an adaptation of Adam that uses a more accurate \emph{examplewise variance estimate} of the gradient. Our preliminary experiments show it can lead to improved performance over Adam in some cases.

These results for the Eve algorithm suggest a more general insight, that when designing gradient-based optimizers for machine learning, one should assume access not to a single gradient estimate per iteration, but to a batch of i.i.d.\ gradient estimates. This better reflects the information that is actually available, given the common practice of evaluating functions on batches of data in parallel, on modern hardware such as graphics processing units (GPUs).

Based on this idea, we propose a new framework for designing gradient-based optimization algorithms. We show that arbitrary statistics of examplewise gradients for adaptive optimization, including higher-order moments, the median and mean absolute deviation, can be obtained using standard software tools. 
In contrast, previous work such as BackPACK \cite{dangel2020backpack} relies on a specialized algorithm \cite{balles2017coupling, goodfellow2015efficient} that can efficiently obtain statistics derived from the examplewise second moment.
We provide a model-agnostic example implementation that supports any differentiable network architecture, including residual \cite{he2016deep}, recurrent \cite{hochreiter1997long} and attention-based \cite{vaswani2017attention} networks. 
This is in contrast to BackPACK, which requires custom code per operation type and only supports chains of linear layers with activations, without fan-in or fan-out.

On a perfectly parallel machine, computing other statistics in addition to the mean should incur minimal per-iteration runtime overhead for deep enough models. We find that in practice runtime overhead varies depending on the architecture. We demonstrate the feasibility of our more general approach by testing Eve on large-scale problems but leave wider exploration of this new optimizer design space for future work.

\section{Examplewise Variance Estimation (Eve)}

In this section, we motivate and derive Eve, a first-order optimizer based on Adam \cite{kingma2014adam}, making use of examplewise gradients. For non-batched optimization ($B=1$), Eve is equivalent to Adam, inheriting its convergence guarantees. While not evaluated in this paper, it is straightforward to transfer the idea of Eve to other adaptive optimizers based on batch gradient second moments, such as Adadelta \cite{zeiler2012adadelta} and Rmsprop \cite{graves2013generating}.

Adam computes exponential moving average statistics $\m$ and $\sq$ for the first and second moment $\mathbb{E} \batchg$ and $\mathbb{E} \batchg^2$ of the batch mean gradient $\batchg$. At each timestep, parameters are updated via
\begin{equation}
\param \gets \param -\alpha\frac{\m}{\sqrt{\sq}+\epsilon},
\end{equation}
where $\epsilon$ is a hyperparameter which is usually small and intended to improve numerical stability. Ignoring $\epsilon$, the update can be rewritten as
\begin{equation}
\param \gets \param -\text{sign}(\m) \frac{\alpha}{\sqrt{\brel^2+1}},
\end{equation}
where $\brel:=\sqrt{\frac{\sq-\m^2}{\m^2}}$
is a statistic for the batch gradient's relative standard deviation $\frac{\sqrt{\var \p{\batchg}}}{|\mathbb{E} \batchg|}$ \cite{balles2018dissecting}.

We therefore propose to use an exponential moving average statistic $\sqel$ of the second moment $\mathbb{E} \g^2$ of the examplewise gradient $\g$ to obtain a more accurate batch gradient variance estimate.
Due to the variance sum law, batch gradient variance $\var \p{\batchg}$ is related to the examplewise variance $\var \p{\g}$ via $\var \p{\batchg} = \frac{1}{\B} \var \p{\g}$, resulting in
\begin{equation}
\frac{\sqrt{\var \p{\batchg}}}{|\mathbb{E} \batchg|} = \frac{\sqrt{\frac{1}{\B} \var \p{\g}}}{|\mathbb{E} \batchg|}.
\end{equation}

This motivates the use of $\rel := \sqrt{\frac{1}{\B} \frac{\sqel-\m^2}{\m^2}}$ instead of $\brel$ to estimate the relative batch gradient standard deviation. For a normally distributed, slowly-changing gradient distribution, the standard deviation of the estimator $\rel$ (`standard error') is approximately \(\sqrt{B}\) lower than $\brel$'s (derivation in \cref{sec:stderror}), demonstrating its improved stability. Using $\rel$ results in the following update for Eve:

\begin{equation}
\param \gets \param -\alpha\frac{\m}{\sqrt{\frac{1}{\B} \sqel + \p{1 - \frac{1}{\B}} \m^2} + \epsilon}.
\end{equation}

The expression $\frac{1}{\B} \sqel + \p{1 - \frac{1}{\B}} \m^2$ is a second moment statistic of the batch gradient $\batchg$. \Cref{alg:adameve} shows the full pseudo-code for the Eve algorithm.
\begin{algorithm2e}
  \caption{Eve optimizer. Differences from Adam are highlighted: Eve becomes equivalent to Adam when using $\textcolor{blue}{\p{\frac{1}{B}\sum_{\batchi=1}^{B} \g_\batchi}^2}$ instead of $\textcolor{blue}{\frac{1}{B}\sum_{\batchi=1}^{B} \g_\batchi^2}$ in line 10, and omitting line 13. As discussed in \cref{sec:examplewise}, lines 7 to 14 can be computed layerwise to lower runtime and peak memory. This can be implemented by leveraging compiler optimization and is not shown here for simplicity.}\label{alg:adameve}
  \SetAlgoNoEnd \SetAlgoNoLine \LinesNumbered \SetKwInput{Require}{Require}
  
  \Require{$\alpha$: Stepsize}
  \Require{$\beta_1, \beta_2 \in [0, 1)$: Exponential decay rates for the moment estimates}
  \Require{$\objective$: Objective function with parameters $\param$}
  \Require{$batch$: Function returning a batch of training data for a given time $t$}
  \Require{$\param$: Initial parameter vector}
  $\biasm \gets 0$ (Initialize first moment vector) \\
  $\biassq \gets 0$ (Initialize second moment vector) \\
  $t \gets 0$ (Initialize timestep) \\
  \While{$\param$ not converged}{
    $t \gets t+1$ \\
    $\data_{1..\B} \gets batch(t)$ (Get next data batch) \\
    \For{$\batchi \gets 1..\B$}{
        $\g_\batchi \gets \nabla_\param \objective_\param \p{\data_\batchi}$ (Get examplewise gradient of objective)
    }
    $\biasm \gets \beta_1 \cdot \biasm + (1 - \beta_1) \cdot \frac{1}{\B}\sum_{\batchi=1}^{\B} \g_\batchi$ (Update biased \first moment estimate) \\
    $\biassq \gets \beta_1 \cdot \biassq + (1 - \beta_1) \cdot \textcolor{blue}{\frac{1}{\B}\sum_{\batchi=1}^{\B} \g_\batchi^2}$ (Update biased examplewise \second moment estimate) \\
    $\m \gets \biasm / \p{1-\beta_1^t}$ (Compute \first moment estimate) \\
    $\sq \gets \biassq / \p{1-\beta_2^t}$ (Compute examplewise \second moment estimate, called $\sqel$ in the text) \\
    $\textcolor{blue}{\sq \gets \frac{1}{\B} \sq + \p{1 - \frac{1}{\B}} \m^2}$ (Compute batch \second moment estimate) \\
    $\param \gets \param -\alpha \m / \p{\sqrt{\sq}+\epsilon}$ (Update parameters)
  } 
  \Return{$\param$}
\end{algorithm2e}

\section{Evaluating Examplewise Gradients} \label{sec:examplewise}
The function \(\batchg\) can usually be computed using automatic differentiation software. However, the sum on the right-hand side of \cref{eq:g-hat} is typically computed internally without exposing the summands to the user. One convenient way to access the `examplewise gradient' terms \(\grad_\param \objective(x_{d_b}, \param)\) is to leverage a vectorizing map function.
Assuming the cost function $f$, evaluated on a single element \(\data\) and the parameters $\theta$, has been defined, we can transform it by taking the gradient w.r.t.\ $\theta$ and applying a vectorized mapping over $x$. 
An example implementation using JAX is shown in \cref{listing:grads}.
It is transferable to other AD libraries with support for compilation and automatic vectorization, such as PyTorch \cite{paszke2019} and TensorFlow \cite{agarwal2019auto}.

\begin{lstlisting}[caption={A training step implementation which uses JAX to evaluate the gradient of a function \texttt{f} on each element of a batch of data \texttt{xs}. The batch of gradient values in \texttt{grads} is used to update the parameters \texttt{theta}. The \texttt{jax.jit} decorator means that the \texttt{training\_step} function will be just-in-time compiled the first time that it is called.},label={listing:grads}]
def f(x, theta):
    # <cost function for a single datapoint x>

@jax.jit
def training_step(xs, theta, opt_state):
    # Differentiate f w.r.t. the argument in position 1,
    # and create a closure over theta:
    grad_elem_f = lambda x: jax.grad(f, 1)(x, theta)
    
    # Evaluate the batch of gradients on the batch of data xs:
    grads = jax.vmap(grad_elem_f)(xs)
    
    # Use an optimizer update rule to update theta
    return optimizer.apply_gradients(opt_state, theta, grads)
\end{lstlisting}

\keypoint{Memory and runtime.} \citet{balles2017coupling} correctly note that storing $B$ examplewise gradients in memory may be infeasible if the number of parameters is high. However, in our method, examplewise gradients for different layers never have to be in memory \textit{at the same time}. Instead, they can be computed in sequence as part of backpropagation, and discarded once the desired statistics are extracted for each layer. For a sequential network with $N$ parameters per layer, this results in additional peak memory usage of $O(BN)$, independent of the layer count.
In our implementation, optimizations of the execution order of this type are automatically performed by the \lstinline{jit} compiler.
Our approach also exploits that all additional examplewise operations are not on the critical path of backpropagation, except for the first layer. The additional required runtime on parallel hardware with enough memory and processors is therefore constant w.r.t. the depth of the network, becoming insignificant for deep enough models. We found that in practice our implementation does have non-negligible runtime overhead for some tasks, details are in \cref{sec:tasks}.

\section{Experiments}

In this section, we demonstrate the practicality of our proposed method to obtain examplewise gradients. We also give preliminary results comparing the performance of Eve to Adam.

\keypoint{Eve performs similarly to Adam on various tasks without retuning hyperparameters.} We evaluate Eve and Adam on five tasks: CIFAR-10 image classification \cite{krizhevsky2009learning} with CNNs, addition with an LSTM \cite{hochreiter1997long} sequence-to-sequence model, transformer-based WMT machine translation \cite{vaswani2017attention}, Atari Pong reinforcement learning based on PPO \cite{schulman2017proximal}, and image modelling with PixelCNN++ \cite{salimans2017pixelcnn}. We provide details in \cref{sec:tasks} and full source code\footnote{Code is available at \url{https://github.com/juliuskunze/eve}.}. Preliminary results based on hyperparameters tuned for Adam are shown in \cref{tab:results}. We leave retuning Eve's hyperparameters for future work.

\begin{table}[h!]
\begin{center}
\begin{tabular}{c c c c c c} 
  & CIFAR-10 acc. & Seq2Seq acc. & WMT acc. & Pong reward & PixelCNN log prob.  \\
 \hline
 Eve (ours) & $69.0\pm1.0\%$ & $98.4\pm 0.0\%$ & $70.8\pm 0.04\%$ & $20.8\pm 0.2$ & $-2.948$ \\ 
 Adam & $68.0\pm0.5\%$ & $98.4\pm 0.0\%$ & $71.4\pm 0.11\%$ & $20.5\pm 0.4$ & $-2.950$ \\
\end{tabular}
\caption{\label{tab:results} Evaluation performance on various tasks. Higher is better. Averages and standard deviation over three runs are shown, except for PixelCNN with a single run.}
\end{center}
\end{table}

\begin{wraptable}{r}{0.5\textwidth}
\vspace{-45pt}
\begin{center}
\begin{tabular}{c c c c c c} 
  Batch size & 8 & 32 & 128 & 512  \\
 \hline
 Eve & 64.8\% & 71.3\% & 69.0\% & 69.4\% \\ 
 Adam & 65.9\% & 72.0\% & 68.0\% & 66.2\% \\
\end{tabular}
\caption{\label{tab:batchsize} Evaluation accuracy for various batch sizes on CIFAR-10 based on single runs.}
\end{center}
\end{wraptable}

\keypoint{Performance gain increase with larger batch size.} 
\Cref{tab:batchsize} shows results for different batch sizes on CIFAR-10. While Adam performs better for smaller batches, Eve has an advantage over Adam for larger batch sizes. 

\begin{wrapfigure}{r}{0.4\textwidth}
    \vspace{10pt}
    \centering
    \includegraphics[width=0.4\textwidth]{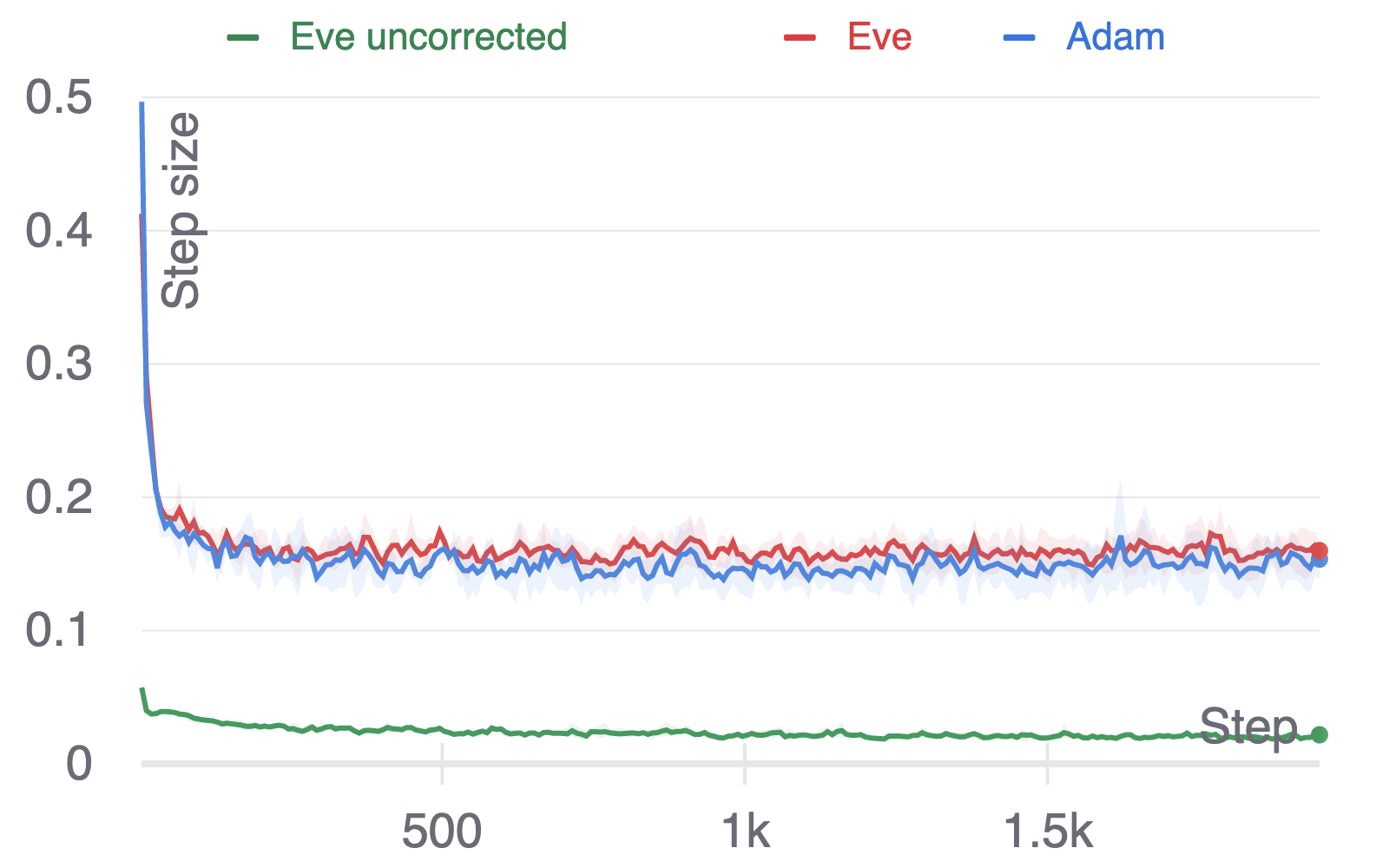}
    \caption{\label{fig:stepsize} Parameter update step length over time on CIFAR-10.}
\end{wrapfigure}

\keypoint{Estimating examplewise variance only is insufficient.} 
We evaluate Eve on CIFAR-10 without the batch variance correction on line 13 in \cref{alg:adameve}. Evaluation accuracy drops drastically to 55.6\%. Comparing the step size length for this ablation, Eve and Adam (\cref{fig:stepsize}) shows that step sizes are much smaller, indicating the importance of the correction term. 

\keypoint{Eve's step size is similar to Adam's but is more stable.} Eve's standard deviation of the step size is 25\% lower on CIFAR-10. Noise from stochastic gradients can help to escape bad local minima \cite{kleinberg2018alternative, zhu2018anisotropic}. Since Adam's less stable variance estimates translate into noisy step sizes, this might explain the performance lead of Adam over Eve on WMT.
Further work is needed to test this hypothesis and understand when such noise is beneficial or harmful to learning.

\keypoint{Runtime and memory usage are practical even for large models.} We report relative runtimes in section \ref{sec:tasks}. There is virtually no slowdown of Eve compared to Adam on the Seq2Seq tasks, which is expected due to the long critical path. Our experiments confirm that our approach is feasible in practice even for large models and batch sizes, such as WMT with 200M parameters and $B=256$.

\bibliography{main}
\newpage
\clearpage
\appendix

\section{Standard Error of Standard Deviation Estimator in Adam vs. Eve} \label{sec:stderror}

The standard deviation (`standard error') of the population variance estimator $s^2=\frac{1}{N} \sum_{i=1}^N \g_\batchi^2 - \p{\frac{1}{N} \sum_{i=1}^N \g_\batchi}^2$ for $N$ i.i.d. normal samples $\g_1, \ldots, \g_N \sim \gauss{\mu, \sigma^2}$ is \cite[Example 4.4, 4.12]{rossi2018mathematical}:
\begin{equation} \label{eq:sq}
\se \p{s^2} = \frac{N - 1}{N} \sqrt{\frac{2}{N - 1}} \sigma^2 \approx \sqrt{\frac{2}{N}} \sigma^2.
\end{equation}

The analog population variance estimator $\bs^2=\frac{1}{\bn} \sum_{i=1}^{\bn} \batchg_\batchi^2 - \p{\frac{1}{\bn} \sum_{i=1}^{\bn} \batchg_\batchi}^2$ applied to the $ \bn = \frac{N}{B}$ means
$\batchg_1, \ldots, \batchg_{\bn}$ of batches of size $B | N$, with
$\batchg_i = \frac{1}{B} \sum_{\batchi=1}^B \g_{(i-1)B + \batchi} \sim \gauss{\mu, \frac{1}{B}\sigma^2}$, has the standard error
\begin{equation} \label{eq:bsq}
\se \p{\bs^2} 
\approx \sqrt{\frac{2 B}{N}} \frac{1}{B}\sigma^2 
\stackrel{\text{(\ref{eq:sq})}}{\approx} \frac{1}{\sqrt{B}} \se \p{s^2}.
\end{equation}

Using the delta method \cite[pp. 386-391]{rao1973linear}, we obtain an approximation of the standard error of the batch mean standard deviation estimators $\bs$ and $\frac{1}{\sqrt{B}} s$ \cite[Example 4.18]{rossi2018mathematical}:
\begin{equation} \label{eq:s}
\se\p{\frac{1}{\sqrt{B}} s}
= \frac{1}{\sqrt{B}} \se\p{s}
\approx \frac{1}{2 \sigma \sqrt{B}} \se\p{s^2} 
\end{equation}
\begin{equation} \label{eq:bs}
\se \p{\bs} \approx \frac{\sqrt{B}}{2 \sigma} \se\p{\bs^2}
\stackrel{\text{(\ref{eq:bsq})}}{\approx} \frac{1}{2 \sigma} \se\p{s^2}
\stackrel{\text{(\ref{eq:s})}}{\approx} \sqrt{B} \se\p{\frac{1}{\sqrt{B}} s}.
\end{equation}
Transferring this analysis to Adam and Eve, $\brel$ and $\rel$ have a relationship analogous to $\bs$ and $\frac{1}{\sqrt{B}} s$, assuming a normal and slowly-changing gradient distribution. Then,
\begin{equation}
\se \p{\rel} 
\stackrel{\text{(\ref{eq:bs})}}{\approx} \frac{1}{\sqrt{B}} \se \p{\brel},
\end{equation}
so we expect that Eve's $\rel$ standard deviation estimator is more stable than Adam's $\brel$ by a factor of $\sqrt{B}$.

\pagebreak
\section{Task details} \label{sec:tasks}

\begin{table}[h!]
\begin{center}
\begin{tabular}{lccccc} 
  & CIFAR-10 & Seq2Seq & WMT & Pong & PixelCNN  \\
 \hline
 Batch size & 128 & 128 & 256 & 256 & 320 \\
 Update steps & 7k & 2k & 100k & 400M & 300k \\
 Parameters & 1.1M & 2.2M & 209M & 4M & 49M \\
 Matmuls with parameters along critical path & 4 & 512\textsuperscript{A} & 37\textsuperscript{B} & 6 & 36 \\
 Runtime of Eve relative to Adam\textsuperscript{C} & 120\% & 100\% & 129\% & 118\% & 454\%\textsuperscript{D} \\
\end{tabular}
\caption{\label{tab:tasks} Task details. Notes: \textsuperscript{A}Due to LSTM encoder and decoder rolled out for 256 steps each. \textsuperscript{B}Only matrix multiplications with parameters are counted, not self-attention matrix multiplications. \textsuperscript{C}Implemented in JAX, run on a TPU v3-8. \textsuperscript{D}We suspect that this massive slowdown is caused by a bug.}
\end{center}
\end{table}

\end{document}

%% file: commands.tex
\newcommand \p[1] {\left(#1\right)}

\newcommand \var{\mathrm{Var}}

\newcommand \gauss[1] {\operatorname{\mathcal{N}}\p{#1}}

\newcommand{\param}{\theta}
\newcommand{\g}{g}
\newcommand{\batchg}{\overline{g}}
\newcommand{\batchi}{b}
\newcommand{\data}{x}

\newcommand{\bs}{\overline{s}}
\newcommand{\bn}{\overline{N}}

\newcommand{\brel}{\overline{\eta}}
\newcommand{\rel}{\eta}

\newcommand{\debias}[1]{\widehat{#1}}
\newcommand{\biasm}{m}
\newcommand{\m}{\debias{\biasm}}
\newcommand{\biassq}{v}
\newcommand{\sq}{\debias{\biassq}}
\newcommand{\sqel}{\sq_e}
\newcommand{\objective}{f}

\newcommand{\B}{B}

\newcommand{\first}{1\textsuperscript{st} }
\newcommand{\second}{2\textsuperscript{nd} }

\DeclareMathOperator*{\argmin}{arg\,min}
\newcommand\grad{\nabla}

\newcommand{\keypoint}[1]{\textbf{#1}}

\newcommand{\se}{\textrm{SE}}

%% file: main.bbl
\begin{thebibliography}{20}
\providecommand{\natexlab}[1]{#1}
\providecommand{\url}[1]{\texttt{#1}}
\expandafter\ifx\csname urlstyle\endcsname\relax
  \providecommand{\doi}[1]{doi: #1}\else
  \providecommand{\doi}{doi: \begingroup \urlstyle{rm}\Url}\fi

\bibitem[Agarwal and Ganichev(2019)]{agarwal2019auto}
Ashish Agarwal and Igor Ganichev.
\newblock Auto-vectorizing {TensorFlow} graphs: Jacobians, auto-batching and
  beyond.
\newblock \emph{arXiv preprint arXiv:1903.04243}, 2019.

\bibitem[Balles and Hennig(2018)]{balles2018dissecting}
L.~Balles and P.~Hennig.
\newblock Dissecting {Adam}: The sign, magnitude and variance of stochastic
  gradients.
\newblock In \emph{ICML}, 2018.

\bibitem[Balles et~al.(2017)Balles, Romero, and Hennig]{balles2017coupling}
L.~Balles, J.~Romero, and P.~Hennig.
\newblock Coupling adaptive batch sizes with learning rates.
\newblock In \emph{Conference on Uncertainty in Artificial Intelligence (UAI)},
  2017.

\bibitem[Dangel et~al.(2020)Dangel, Kunstner, and Hennig]{dangel2020backpack}
Felix Dangel, Frederik Kunstner, and Philipp Hennig.
\newblock Back{PACK}: Packing more into backprop.
\newblock In \emph{ICLR}, 2020.

\bibitem[Goodfellow(2015)]{goodfellow2015efficient}
Ian Goodfellow.
\newblock Efficient per-example gradient computations.
\newblock \emph{arXiv preprint arXiv:1510.01799}, 2015.

\bibitem[Graves(2013)]{graves2013generating}
Alex Graves.
\newblock Generating sequences with recurrent neural networks.
\newblock \emph{arXiv preprint arXiv:1308.0850}, 2013.

\bibitem[He et~al.(2016)He, Zhang, Ren, and Sun]{he2016deep}
Kaiming He, Xiangyu Zhang, Shaoqing Ren, and Jian Sun.
\newblock Deep residual learning for image recognition.
\newblock In \emph{IEEE conference on computer vision and pattern recognition},
  2016.

\bibitem[Hochreiter and Schmidhuber(1997)]{hochreiter1997long}
Sepp Hochreiter and J{\"u}rgen Schmidhuber.
\newblock Long short-term memory.
\newblock \emph{Neural computation}, 9\penalty0 (8), 1997.

\bibitem[Kingma and Ba(2015)]{kingma2014adam}
Diederik~P. Kingma and Jimmy Ba.
\newblock Adam: {A} method for stochastic optimization.
\newblock In \emph{ICLR}, 2015.

\bibitem[Kleinberg et~al.(2018)Kleinberg, Li, and
  Yuan]{kleinberg2018alternative}
Bobby Kleinberg, Yuanzhi Li, and Yang Yuan.
\newblock An alternative view: When does sgd escape local minima?
\newblock In \emph{ICML}, 2018.

\bibitem[Krizhevsky et~al.(2009)Krizhevsky, Hinton,
  et~al.]{krizhevsky2009learning}
Alex Krizhevsky, Geoffrey Hinton, et~al.
\newblock Learning multiple layers of features from tiny images.
\newblock 2009.

\bibitem[Liu et~al.(2020)Liu, Jiang, He, Chen, Liu, Gao, and
  Han]{liu2019variance}
Liyuan Liu, Haoming Jiang, Pengcheng He, Weizhu Chen, Xiaodong Liu, Jianfeng
  Gao, and Jiawei Han.
\newblock On the variance of the adaptive learning rate and beyond.
\newblock In \emph{ICLR}, April 2020.

\bibitem[Paszke et~al.(2019)Paszke, Gross, Massa, Lerer, Bradbury, Chanan,
  Killeen, Lin, Gimelshein, Antiga, Desmaison, Kopf, Yang, DeVito, Raison,
  Tejani, Chilamkurthy, Steiner, Fang, Bai, and Chintala]{paszke2019}
Adam Paszke, Sam Gross, Francisco Massa, Adam Lerer, James Bradbury, Gregory
  Chanan, Trevor Killeen, Zeming Lin, Natalia Gimelshein, Luca Antiga, Alban
  Desmaison, Andreas Kopf, Edward Yang, Zachary DeVito, Martin Raison, Alykhan
  Tejani, Sasank Chilamkurthy, Benoit Steiner, Lu~Fang, Junjie Bai, and Soumith
  Chintala.
\newblock Pytorch: An imperative style, high-performance deep learning library.
\newblock In \emph{NeurIPS}. 2019.

\bibitem[Rao(1973)]{rao1973linear}
Calyampudi~Radhakrishna Rao.
\newblock \emph{Linear statistical inference and its applications}, volume~2.
\newblock Wiley New York, 1973.

\bibitem[Rossi(2018)]{rossi2018mathematical}
Richard~J Rossi.
\newblock \emph{Mathematical statistics: An introduction to likelihood based
  inference}.
\newblock John Wiley \& Sons, 2018.

\bibitem[Salimans et~al.(2017)Salimans, Karpathy, Chen, and
  Kingma]{salimans2017pixelcnn}
Tim Salimans, Andrej Karpathy, Xi~Chen, and Diederik~P. Kingma.
\newblock {PixelCNN++}: A {PixelCNN} implementation with discretized logistic
  mixture likelihood and other modifications.
\newblock In \emph{ICLR}, 2017.

\bibitem[Schulman et~al.(2017)Schulman, Wolski, Dhariwal, Radford, and
  Klimov]{schulman2017proximal}
John Schulman, Filip Wolski, Prafulla Dhariwal, Alec Radford, and Oleg Klimov.
\newblock Proximal policy optimization algorithms.
\newblock \emph{arXiv preprint arXiv:1707.06347}, 2017.

\bibitem[Vaswani et~al.(2017)Vaswani, Shazeer, Parmar, Uszkoreit, Jones, Gomez,
  Kaiser, and Polosukhin]{vaswani2017attention}
Ashish Vaswani, Noam Shazeer, Niki Parmar, Jakob Uszkoreit, Llion Jones,
  Aidan~N Gomez, {\L}ukasz Kaiser, and Illia Polosukhin.
\newblock Attention is all you need.
\newblock In \emph{NeurIPS}, 2017.

\bibitem[Zeiler(2012)]{zeiler2012adadelta}
Matthew~D Zeiler.
\newblock Adadelta: an adaptive learning rate method.
\newblock \emph{arXiv preprint arXiv:1212.5701}, 2012.

\bibitem[Zhu et~al.(2018)Zhu, Wu, Yu, Wu, and Ma]{zhu2018anisotropic}
Zhanxing Zhu, Jingfeng Wu, Bing Yu, Lei Wu, and Jinwen Ma.
\newblock The anisotropic noise in stochastic gradient descent: Its behavior of
  escaping from sharp minima and regularization effects.
\newblock \emph{arXiv preprint arXiv:1803.00195}, 2018.

\end{thebibliography}
